# Imbalanced Learning-Based Automatic SAR Images Change Detection by Morphologically Supervised PCA-Net

Rongfang Wang, *Member, IEEE*, Jie Zhang, Jiawei Chen, *Member, IEEE*, Licheng Jiao, *Fellow, IEEE*, and Mi Wang

*Abstract*—Change detection is a quite challenging task due to the imbalance between unchanged and changed class. In addition, the traditional difference map generated by log-ratio is subject to the speckle, which will reduce the accuracy. In this letter, an imbalanced learning-based change detection is proposed based on PCA network (PCA-Net), where a supervised PCA-Net is designed to obtain the robust features directly from given multitemporal synthetic aperture radar (SAR) images instead of a difference map. Furthermore, to tackle with the imbalance between changed and unchanged classes, we propose a morphologically supervised learning method, where the knowledge in the pixels near the boundary between two classes is exploited to guide network training. Finally, our proposed PCA-Net can be trained by the data sets with available reference maps and applied to a new data set, which is quite practical in change detection projects. Our proposed method is verified on five sets of multiple temporal SAR images. It is demonstrated from the experiment results that with the knowledge in training samples from the boundary, the learned features benefit change detection and make the proposed method outperform than supervised methods trained by randomly drawing samples.

## I. Introduction

CHANGE detection indicates the process of identifying differences in the state of an object or phenomenon by observing it at different times [1], and it has been widely applied in the evaluation on the attack effect in combatant and the survey of the geographical information in a certain region. In the past decades, remote sensing images have been widely used for change detection.

Manuscript received March 7, 2018; revised June 5, 2018 and October 5, 2018; accepted October 24, 2018. Date of publication November 13, 2018; date of current version March 25, 2019. This work was supported in part by the National Natural Science Foundation of China under Grant 61701361, in part by the Open Fund of State Laboratory of Information Engineering in Surveying, Mapping and Remote Sensing, Wuhan University, under Grant 17E02, in part by the Key R&D Program, The Key Industry Innovation Chain of Shaanxi, under Grant 2018JM6083, and in part by the Fundamental Research Funds for the Central University under Grant JB181701. *(Corresponding author: Jiawei Chen.)*

R. Wang, J. Zhang, J. Chen, and L. Jiao are with the Key Laboratory of Intelligent Perception and Image Understanding, Ministry of Education, School of Artificial Intelligence, Xidian University, Xi'an 710071, China (e-mail: rfwang@xidian.edu.cn; jawaechan@gmail.com).

M. Wang is with the State Key Laboratory of Information Engineering in Surveying, Mapping and Remote Sensing, Wuhan University, Wuhan 430079, China.

Color versions of one or more of the figures in this letter are available online at http://ieeexplore.ieee.org.

Digital Object Identifier 10.1109/LGRS.2018.2878420

Synthetic aperture radar (SAR) is a sensor working without the limitations of seasons and weather. It transmits microwaves to the surface of the earth and receives the echo data reflected from ground, and a set of SAR image is then generated from the echo data. Therefore, it contains much useful information about the terrain of the earth. Therefore, SAR images have been popularly employed in change detection due to many advantages of SAR.

Bruzzone and Prieto [2], [3] proposed a change detection based on difference map, where changed regions are detected based on the analysis of difference map. However, an SAR image is usually contaminated by the speckle, and the simple difference map based on minus operator will be inaccurate. To tackle this issue, Bovolo and Bruzzone [4] proposed a difference map based on log-ratio operator according to the fact that the speckle is modeled as multiplicative noise. Gong *et al.* [5] proposed an improved log-ratio operator by the spatial context information, which is more robust to the speckle. Although there are many other improvement methods [6]–[9] proposed for change detection, the traditional difference maps based on minus or log-ratio operators are subject to the speckle. The difference map with noisy contamination will reduce the accuracy of change detection.

To solve this issue, some methods were proposed to reduce the influence of the speckle. Li *et al.* [10] proposed a pairwise dictionary learning method to obtain robust features from difference maps. Recently, deep learning methods have become increasingly popular and employed to computer vision tasks. Zhang *et al.* [11] employed the deep representation to hyperspetral image change detection. Gong *et al.* [12] proposed a change detection method based on the deep neural networks (DNNs). Liu *et al.* [13] proposed a deep convolutional couple network for multisource remote sensing image change detection. However, most DNNs are time-consuming. Chan *et al.* [14] proposed PCA network (PCA-Net) for image classification, and it is considered as a simplified DNN. Gao *et al.* [15] applied the PCA-Net to SAR image change detection.

However, most of the above-mentioned methods are unsupervised, and few works focus on supervised learning-based SAR image change detection. In fact, the information from the reference map is quite important to guide the change detection. It is demonstrated that supervised information can improve the performance of neural networks [16]. Liu *et al.* [17] proposed a stacked restricted Boltzmann machines method for change detection, where supervised information is employed for network parameter fine-tuning. However, changed regions usually occupy much smaller proportion, and unchanged regions will dominate detection result, which is considered as a typical imbalanced classification in the context of machine learning. Some methods have been proposed for imbalanced classification [18], and it is important to select moderate number of training samples, which is ignored by most change detection methods.

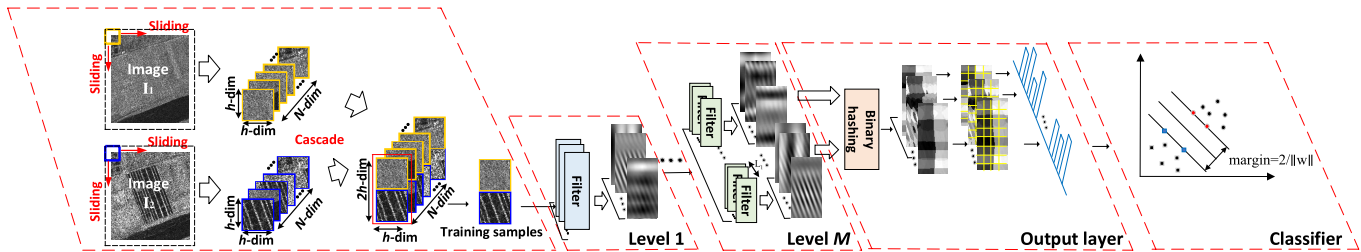

Fig. 1. Framework of SAR image change detection based on PCA-Net.

Inspired by this, in this letter, an imbalanced learning-based method is proposed for SAR image change detection. In the proposed method, a supervised PCA-Net is developed to directly learn robust features from multitemporal SAR images, which reduces the effects of the speckle. In the last layer of the network, a linear classifier is appended, and it is trained by the learned feature map and corresponding ground truth. Furthermore, to solve the imbalance classification issue, a morphologically supervised learning is proposed, where the samples along the boundary between changed regions and unchanged regions are selected for network training. In addition, the supervised PCA-Net can be trained by the data sets with available reference maps and applied to a new data set, which is practical for the change detection on those data sets without reference maps. The experimental results on five sets of multitemporal SAR images show that the proposed PCA-Net can obtain promising results on these data sets.

The rest of this letter is organized as follows. The proposed algorithm will be introduced in Section II. Experimental results will be presented in Section III, and a short concluding remark will be made in Section IV.

## II. Proposed Method

In this section, an imbalanced learning-based SAR image change detection method will be proposed. First, a supervised PCA-Net method will be introduced, and then, the proposed imbalanced learning method for SAR image change detection will be introduced based on the supervised PCA-Net.

### A. PCA Network for SAR Image Change Detection

The PCA-Net was first proposed [14] for image classification and, then, applied to SAR image change detection [15]. The main difference from convolution neural network is that the filter kernels in the PCA-Net are obtained without back-propagation, which reduces computation complexity. Instead, the filters are considered as the eigenvectors of most large eigenvalues after applying the eigen decomposition to covariance matrix. The whole framework of PCA-Net can be shown as Fig. 1. In addition, being different from traditional framework of SAR image change detection working on noisy difference image, our framework is trained on temporal SAR images to reduce the influence of the noise in a difference map.

Given two temporal SAR images over the same region $I_1$ and $I_2$, suppose they have been aligned. For two patches $s_i^1$ and $s_i^2$ with $h \times h$ from two temporal SAR images at the same location, a training patch $s_i = [s_i^1, s_i^2] \in \mathbb{R}^{2h \times h}$ is constructed by cascading these two patches. The process can be shown in Fig. 1. The goal of PCA-Net is to learn filters, and they will be applied to extract the feature on each patch $s_i$.

To train the filter kernels in the network, another $N$ patches $x_i \in k \times k, i = 1, 2, \ldots, N$ are extracted from training patches $S = \{s_i\}$, and each patch is scaled by removing the mean and normalizing in the unit $\ell_2$-norm space. Then, training set $X \in \mathbb{R}^{k^2 \times N}$ by vectorizing each patch $x_i$. Then, the $M_1$ filter kernels in the first layer can be trained by

$$\min_{U} \|X - UU^T X\|_F^2, \text{ s.t. } U^T U = I_{L_1} \quad (1)$$

where $U = [u_1, u_2, \ldots, u_{L_1}] \in \mathbb{R}^{k^2 \times L_1}$ denotes the $L_1$ eigenvectors of covariance matrix $XX^T$ corresponding the $L_1$ largest eigenvalues. $I_{L_1}$ denotes an identity matrix with size of $L_1$. For each eigenvector $u_i$, the corresponding filter kernels can be obtained by $W_i = \text{mat}(u_i), i = 1, 2, \ldots, L_1$, where $\text{mat}(u)$ denote the operator that reshapes a vector $u \in \mathbb{R}^{k^2}$ into a matrix with $k \times k$. Finally, the output of this layer can be obtained by

$$s_i^j = s_i * W_j, (i = 1, 2, \ldots, N, \ j = 1, 2, \ldots, L_1). \quad (2)$$

The $i$th filtered patch by the $j$th filter $s_i^j$ captures the directional information of the patch $s_i$, and it will be considered as the input of the next layer after normalization. If the network contains $M$ levels, there will be $L_1 L_2 \ldots L_M$ outputs.

In this letter, a two-layer PCA-Net is constructed, and in the second layer, $L_1 L_2$ outputs can be obtained. Following the outputs of the second layer, hashing and histogram [14] are applied to obtain the final feature $f_i$ for each patch $x_i$. Finally, the classifier in the last layer can be trained by the feature $f_i$ of each patch and its label.

### B. Morphologically Supervised Learning

In Section II-A, a PCA-Net has been introduced. In the last layer of the network, a support vector machine (SVM) is employed as a classifier. Gao *et al.* [15] employed highly confident pseudolabels generated by clustering to train the classifier. However, change detection is a typical imbalanced classification task since in most cases, changed region usually accounts a much smaller proportion. It is difficult to obtain as many training samples of changed class as unchanged one. In addition, the pseudolabels generated by clustering are inaccurate, and the unavoidable errors of labels will reduce the performance of the network. Especially, the pixels near the boundary between unchanged and changed regions are difficult to distinct due to the interclass similarity of their intensities.

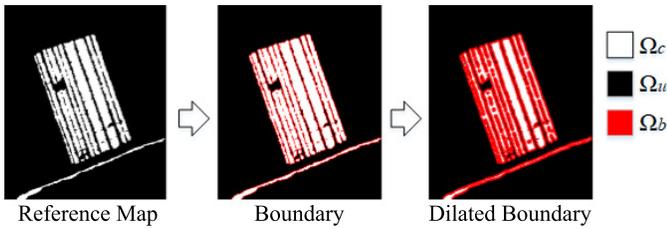

Fig. 2. Morphology-based sample selection.

In fact, a reference image can be divided into boundary set $\Omega_b$, inner changed set $\Omega_c$, and inner unchanged set $\Omega_u$, as shown in Fig. 2, where $\Omega_c$ and $\Omega_u$ only contain changed and unchanged pixels, respectively. Instead, $\Omega_b$ contains both changed pixels and unchanged pixels, which are easy to confuse due to the similarity of their intensity. In most of the earlier methods, it is difficult to obtain accurate classification on these pixels.

In this section, a morphologically supervised learning method is proposed, where the pixels near the boundary are taken account into training as many as possible. To achieve this, first, we find the boundary between changed and unchanged regions from reference image and, then, dilate boundary to cover the region near the boundary. In the construction of training set, several samples are randomly selected from the boundary set $\Omega_b$, and then, more training samples are obtained from $\Omega_c$ and $\Omega_u$ by an oversampling to keep the same number of training samples from the changed and unchanged class.

Being different from most of the earlier methods, in our proposed method, the samples from the boundary set $\Omega_b$ are selected for training, which will effectively improve the performance of classifier.

## III. Experimental Results and Analysis

### A. Experiment Configures

In this section, to verify the advantage of our proposed methods, the comparison experiments are performed on five data sets: Bern, San Fransisco [15], two scenes of Yellow River [12] (YR-A and YR-B), and Ottawa.

The proposed method (PCA-Net-buc) is compared with the PCA-Net trained by pseudolabels [15] (PCA-Net) and the supervised PCA-Net trained with the samples only randomly drawn from changed and unchanged regions (PCA-Net-uc). In addition, we also compared the proposed method with the DNN [12]. It is noted that in above-mentioned four compared methods, the drawn training samples between two classes are imbalanced. In addition, an oversampling strategy will be applied to PCA-Net-buc to keep the balance of training samples between two classes, which is shortened as PCA-Net-obuc. The comparison experiments are conducted on PC with 2.2 GHz CPU, 16-GB RAM, and the MATLAB 2016 on windows operation system.

In the PCA-Net, two layers are configured, and in each layer, eight filters are employed. In this case, there are 8 and 64 outputs in the first and the second layers, respectively. In the last layer, an SVM classifier is employed to get the final classification results. For the DNN method, the training samples are randomly drawn from unchanged and changed regions, respectively, with the patch size being $5 \times 5$. In addition, the network is trained on a simple difference map proposed in [12].

### B. Comparison Experiment Results

In the comparison experiments, we take the patch size as $7 \times 7$, $11 \times 11$, and $15 \times 15$ and select the optimal patch size for each compared method. Moreover, each method is verified at different proportions of training samples. Then, each compared method with the optimal patch size is run at multiple proportions of training samples, where each case is run for 10 times, and the mean and standard deviation (Std) of kappa coefficient are used to evaluate its performance. The comparison results on those four data sets are shown in Fig. 3.

It can be shown from Fig. 3 that our proposed method shows great advantages over other compared methods. First, the performances of PCA-Net do not increase when the proportion of training sample increases. It indicates that the performance of the model trained by pseudolabels cannot be improved even the available training samples increases. The essential reason is that the inaccurate pseudolabels provide incorrect information in network training. Instead, the performances of PCA-Net-uc improve with the increase of training samples. It means that the accurate labels are important for improving performance. Moreover, when the proportion of training samples is up from 30%, the supervised PCA-Net methods outperform than other compared methods. When the proportion of training sample is below to 30%, DNN performs worse than PCA-Net-uc, even worse than PCA-Net on San Fransisco data set and YR-A data set, while PCA-Net-buc and PCA-Net-obuc still perform better than other compared methods. Furthermore, the performances of PCA-Net-buc and PCA-Net-obuc do not dramatically decrease, even when less training samples are available. Next, the performances of PCA-Net-buc and PCA-Net-obuc are comparable in all proportions of training samples. It indicates that the samples along the boundary are key for the model training, which, almost, can be considered as the support vectors in the context of SVM. Finally, for the YR-A data set with the strong speckle. The performance of most of compared methods decrease dramatically, while the Kappa coefficients of our proposed method can keep being above 0.9 even only 10% training samples are available.

To further show the benefit of our proposed method, the visual results in the extreme case are shown in Fig. 4, where the proportion of training sample equals 5%. Typically, for Bern data set, in both PCA-Net and PCA-Net-uc methods, more details on changed regions are missed. Instead, our method obtains more completed changed region with many details, especially only a few training samples from changed region are used. For YR-A data set, both two temporal images are contaminated by the strong speckles with different levels, which is a great challenge in change detection. It is shown from the third row of Fig. 4 that the changes of river coast are missed with different levels. Instead, this change spot can be almost totally detected by our proposed method.

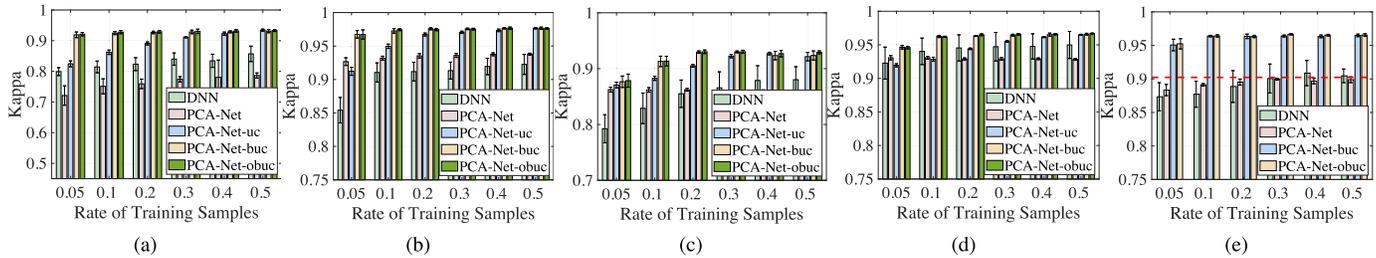

Fig. 3. Evaluations on (a) Bern, (b) San Fransisco, (c) YR-A, (d) Ottawa, and (e) YR-B data sets.

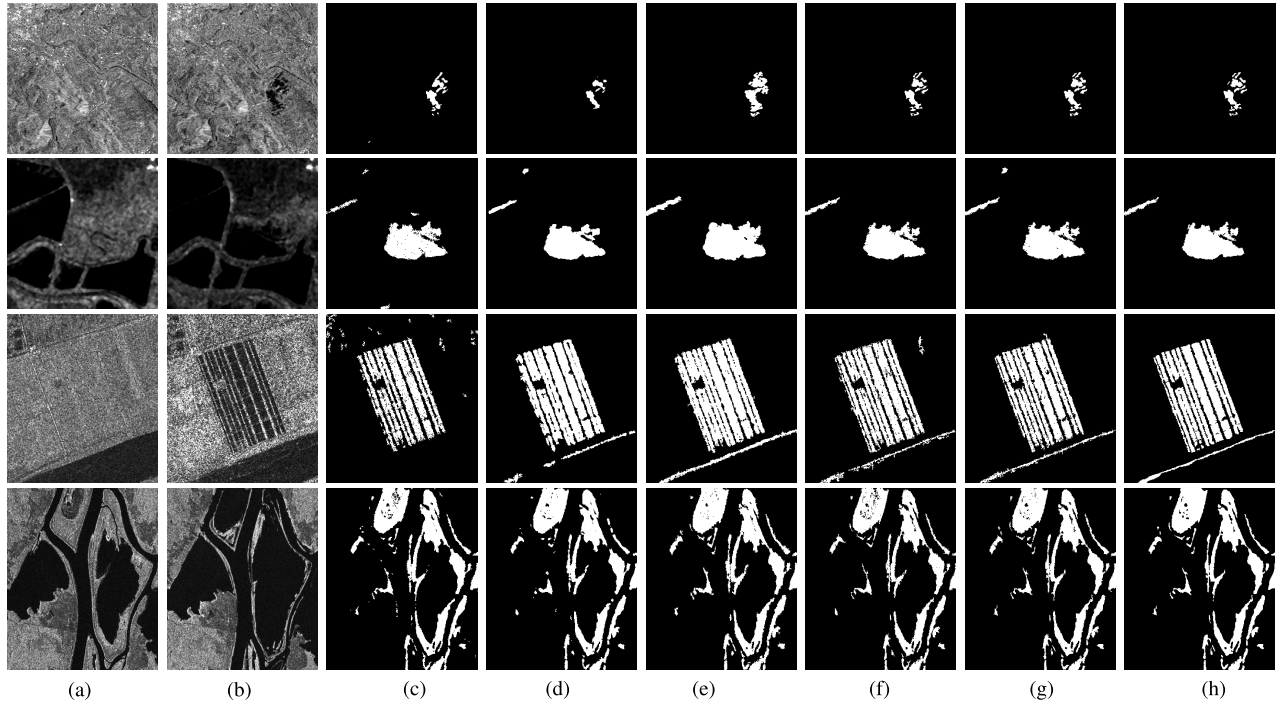

Fig. 4. Visual results of various change detection methods in a low proportion of training samples (5%). (Top) to (Down) Results on Bern, San Fransisco, YR-A, and Ottawa data sets, respectively. (a) Image T1. (b) Image T2. (c) DNN. (d) PCA-Net. (e) PCA-Net-uc. (f) PCA-Net-buc. (g) PCA-Net-obuc. (h) Reference.

Considering all the above-mentioned comparison results, it can be observed that the samples near the boundary are quite important in classifier training. In PCA-Net, the training samples are selected by clustering, and the ones near the clusters are taken for training. However, the selected samples are not necessary near the boundary between changed and unchanged regions, and the pseudolabels from the clustering are incorrect with a certain probability. The model trained by incorrect labels will reduce its performance. In PCA-Net-uc, the training samples are randomly drawn. Even with the ground truth, the training samples are less from the boundary. Especially, when the available training samples are quite few, it will perform badly for those pixels near the boundary between two classes. Instead, in our method PCA-Net-buc, the samples near the boundary and their true labels are selected for training. These easy-to-misclassify samples with their ground truth will

TABLE I
SIGNIFICANCE OF OUR METHOD ($p = 5 \times 10^{-3}$) AT DIFFERENT RATES (PROPORTIONS OF TRAINING SAMPLES)

| rates | 5% | 10% | 20% |
|---|---|---|---|
| test value | -30.62 | -32.79 | -19.60 |
| p-value | $2.06 \times 10^{-10}$ | $1.12 \times 10^{-10}$ | $1.08 \times 10^{-8}$ |
| rates | 30% | 40% | 50% |
| test value | -10.13 | -2.58 | 1.88 |
| p-value | $3.19 \times 10^{-6}$ | 0.03 | 0.09 |

give important guidance for classifier training, even a very few of them are selected. It has been shown from Fig. 3 that supervised PCA-Nets outperform than PCA-Net. To further evaluate the effects of morphologically supervised learning, we evaluate the statistical significance of the performances of PCA-Net-buc over PCA-Net-uc. Here, we apply T-test at a certain level ($p = 5 \times 10^{-3}$). The significance is evaluated at different proportions of training samples. The results are shown in Table I. We recorded the test values and p-values as evaluation, where the larger absolute test value and the smaller p-values will indicate more significant improvement. It can be seen from Table I that PCA-Net-buc obtains significant

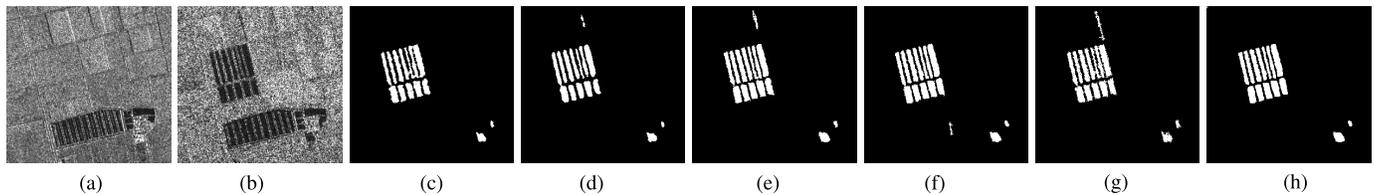

Fig. 5. Visual results on YR-B data set (a) Image T1. (b) Image T2 by (c) DNN, (d) unsupervised PCA-Net, (e) and (f) supervised PCA-Net trained by YR-B, (g) supervised PCA-Net trained by YR-A, and (h) Reference.

improvement over PCA-Net-uc, especially when the rate is lower than 40%.

### C. Generalization of the Proposed Method

To further verify the effect of the proposed method, in this experiment, we train a PCA-Net (PCA-Net-gen) by YR-A and apply the trained network to YR-B. In the training stage, we select all the pixels along the boundaries between changed and unchanged regions and a half-pixel's inner regions as training samples. We compare the kappa in this experiment with the ones in the previous experiments. The visual results are shown in Fig. 5.

In Fig. 5, we compare PCA-Net-gen with DNN, PCA-Net, PCA-Net-buc, and PCA-Net-obuc, which are trained by 5% samples. It is shown that the results of PCA-Net-gen are better than DNN and PCA-Net but a little more FA than PCA-Net-buc and PCA-Net-obuc. More comparisons of evaluations are shown in Fig. 3(e), where the means and Std of DNN, PCA-Net, PCA-Net-buc, and PCA-obuc are compared, and the kappa of PCA-Net-gen is shown by a red baseline. We can observe that the kappa of PCA-Net-gen reaches 0.9, which is little less than PCA-Net-buc and PCA-Net-obuc, but it outperforms than DNN and PCA-Net, when the proportion of training samples is less than 0.3. It is expectable that we will get better results when more data sets are used for training. It means that it is feasible to train the PCA-Net on several available training data sets and obtain the promising results on a new data set. By this manner, people can get the confidential reference map of a new data set with less effort, which is quite practical in change detection projects.

## IV. CONCLUSION

SAR image detection is a typical imbalanced classification task. In this letter, an imbalanced learning method has been developed based on a supervised PCA-Net. To tackle with this imbalance classification task, in the proposed method, we explored the knowledge from those samples near the boundaries between changed and unchanged regions to guide the training of network and classifier. The experimental results demonstrate that the proposed method outperforms than the models trained by randomly drawn samples, especially when the available training samples are quite limited. Finally, the PCA-Net can be trained on the data sets with available reference maps and applied to a new one, which is quite useful in practical change detection projects.